\documentclass[conference]{IEEEtran}
\IEEEoverridecommandlockouts
\usepackage{cite}
\usepackage{amsmath, amssymb, amsfonts}
\usepackage{algorithmic}
\usepackage{graphicx}
\usepackage{textcomp}
\usepackage{xcolor}

\usepackage{bm}
\usepackage{mdframed}
\usepackage{multirow}
\usepackage{graphicx}
\usepackage{subcaption} % For subfigures
\usepackage{soul}
\usepackage{xcolor}

\usepackage{rotating}
\usepackage{hyperref}
\usepackage{booktabs}
\usepackage[most]{tcolorbox}

{\normalsize}
\definecolor{updategreen}{RGB}{0,128,0} % adjust to taste

\def\BibTeX{{\rm B\kern-.05em{\sc i\kern-.025em b}\kern-.08em T\kern-.1667em\lower.7ex\hbox{E}\kern-.125emX}}

\begin{document}
    \title{Reasoning over User Preferences:\\Knowledge Graph‑Augmented LLMs for\\Explainable Conversational Recommendations}

    % \author{\IEEEauthorblockN{Zhangchi Qiu}
    % \IEEEauthorblockA{\textit{School of ICT} \\
    % \textit{Griffith University}\\
    % Gold Coast, Australia \\
    % zhangchi.qiu@griffithuni.edu.au}
    % \and
    % \IEEEauthorblockN{Linhao Luo}
    % \IEEEauthorblockA{\textit{Department of Data Science and AI} \\
    % \textit{Monash University}\\
    % Melbourne, Australia \\
    % linhao.luo@monash.edu}
    % \and
    % \IEEEauthorblockN{Shirui Pan} 
    % \IEEEauthorblockA{\textit{School of ICT} \\
    % \textit{Griffith University}\\
    % Gold Coast, Australia\\
    % s.pan@griffith.edu.au}
    % \and
    % \IEEEauthorblockN{Alan Wee-Chung Liew}
    % \IEEEauthorblockA{\textit{School of ICT} \\
    % \textit{Griffith University}\\
    % Gold Coast, Australia \\
    % a.liew@griffith.edu.au}
    % }
    \author{Zhangchi Qiu$^{1}$, Linhao Luo$^{2}$, Shirui Pan$^{1}$, Alan Wee-Chung Liew$^{1*}$\ \\
    $^{1}$\textit{School of ICT, Griffith University, Gold Coast, Australia} \\
    $^{2}$\textit{Department of Data Science and AI, Monash University, Melbourne, Australia} \\
    zhangchi.qiu@griffithuni.edu.au, linhao.luo@monash.edu, \{s.pan, a.liew\}@griffith.edu.au
    }
    \maketitle
    \begingroup
    \renewcommand\thefootnote{\fnsymbol{footnote}}% 1→*, 2→†, ...
    \footnotetext[1]{Corresponding author.}
    \endgroup
    \begin{abstract}
        Conversational Recommender Systems (CRSs) aim to provide personalized recommendations
        by capturing user preferences through interactive dialogues.
        Explainability in CRSs is crucial as it enables users to understand the
        reasoning behind recommendations, increasing system transparency and trustworthiness.
        However, current CRSs often leverage knowledge graphs (KGs) or language models
        to extract and represent user preferences as latent vectors, which
        limits their explainability. Large language models (LLMs) offer powerful
        reasoning capabilities that can bridge this gap by generating human-understandable
        preference summaries. However, effectively reasoning over user preferences
        in CRSs remains challenging as LLMs pre-trained on large-scale corpora may
        not be well-suited for analyzing user preferences, which requires domain-specific
        knowledge. While KGs provide rich domain knowledge, integrating them
        with LLMs encounters a significant modality gap between structured KG
        information and unstructured conversations.
        In this paper, we propose COMPASS (\textbf{Com}pact \textbf{P}reference \textbf{A}nalyzer
        and \textbf{S}ummarization \textbf{S}ystem), a plug-and-play framework that
        synergizes LLMs and KGs to reason over user preferences, enhancing the
        performance and explainability of existing CRSs. COMPASS employs a two-stage
        training approach: first, it bridges the gap between the structured KG and
        natural language through novel graph entity captioning pre-training.
        This enables the LLM to transform KG entities into natural language
        descriptions, allowing it to comprehend domain-specific knowledge. Next,
        COMPASS optimizes user preference reasoning via knowledge-aware instruction
        fine-tuning, where the LLM learns to reason and summarize user
        preferences from dialogue histories and KG-augmented context. This enables
        COMPASS to perform knowledge-aware reasoning and generate interpretable user
        preferences that can seamlessly integrate with existing CRS models for improving
        recommendation performance and explainability. Our experiments on
        benchmark datasets demonstrate the effectiveness of COMPASS in improving
        various CRS models.
    \end{abstract}


    \section{Introduction}

Providing personalized, context-aware recommendations that align with users' changing
preferences and situational needs is a critical challenge across various domains
such as e-commerce, streaming platforms, and other online services. In recent
years, conversational recommender systems (CRSs) have emerged as a promising approach,
harnessing the power of natural language interactions to unravel user
preferences~\cite{jannach_survey_2021, GAO_advances_2021}. By engaging the user in
interactive conversations, CRSs enable a better understanding of the user's
evolving interests and guide them toward products, services, and information
that best meet their immediate requirements~\cite{christakopoulou_towards_2016, li_towards_2018}.

Despite the success, CRSs face unique challenges in precisely modeling user preferences
from semantically rich and dynamically evolving dialogue. They must interpret
often ambiguous and context-dependent natural language inputs~\cite{li_towards_2018,
zhou_improving_2020}, while continuously updating and refining their understanding
of user preferences in real-time~\cite{sun_conversational_2018}, integrating
immediate interests with overall preferences developed over time.

\begin{figure}[t!]
  \centering
  \includegraphics[width=\columnwidth]{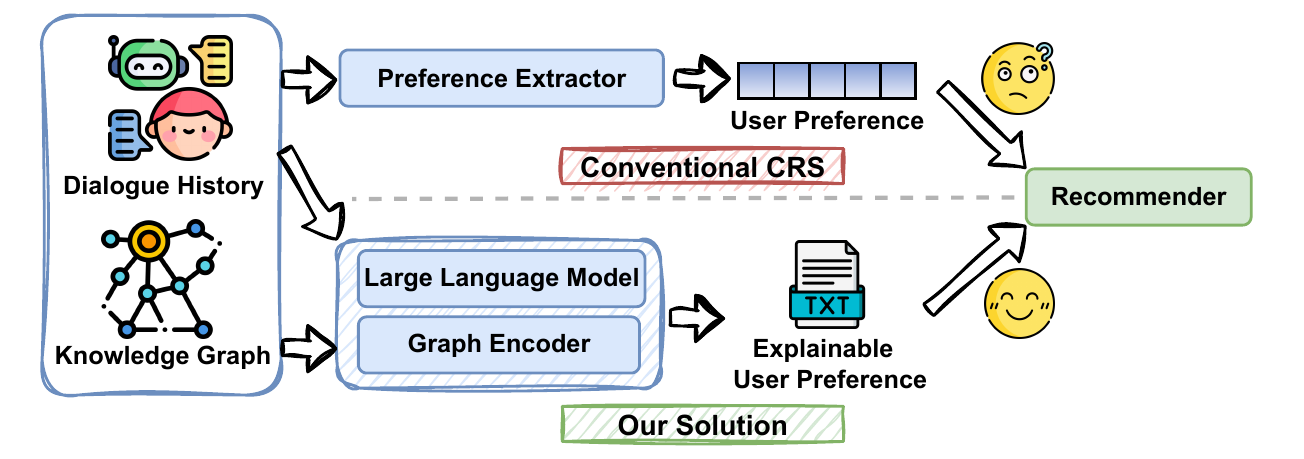}
  % \vspace{-0.2cm}
  \caption{Conventional CRSs lack explainability in user preference modeling by
  extracting hidden representation, while our approach enhances transparency by generating
  interpretable user preference in text.}
  \label{fig:intro_fi}
  \vspace{-0.5cm}
\end{figure}

Conventional CRSs mainly depends on item-centric approaches~\cite{chen_towards_2019,
zhou_improving_2020}, focusing on explicitly mentioned items in the conversation
to model user preference. Recent studies incorporating pre-trained language models
(PLMs) to enhance both natural language understanding and user preference modeling~\cite{wang_towards_2022,
wang-etal-2022-recindial}. However, these methods often fail to infer implicit preferences
or reason about underlying motivations beyond explicit item mentions, leading to
a superficial understanding of user intent. Moreover, they typically represent
user preferences as hidden embeddings, leaving it unclear what specific preference
is being considered when making a recommendation, as shown at the top of Figure~\ref{fig:intro_fi}.
This ambiguity makes it challenging to verify the underlying reasons and results
behind the recommendations, hindering the transparency and accountability of the
recommender system.

Recent advancements in large language models (LLMs), have demonstrated exceptional
capabilities in both language understanding~\cite{zhao2025surveylargelanguagemodels,grattafiori2024llama3herdmodels}
and complex reasoning~\cite{wei_chainofthoughtprompt_2023}. These capabilities
make LLMs promising for explainable recommendations, and recent works~\cite{ma2024xrec,
wang_can_2024,li_g-refer_2025} have demonstrated the LLM's ability to reason over
user historical behaviors and generate interpretable recommendation explanations.
Nevertheless, these methods only focus on the traditional recommendation settings,
where user preferences are static. In contrast, CRSs require more sophisticated
reasoning to model user preferences that evolve dynamically throughout
conversation. While recent work has explored LLMs for various CRS tasks,
including evaluation~\cite{wang-etal-2023-rethinking-evaluation, zhu_how_reliable_2024},
zero-shot recommendation~\cite{he_large_language_2023}, and task planning~\cite{feng_largelanguagemodelenhanced_2023,
fang_multiagentconversa_2024}, the role of explainable user preferences in LLM-based
CRSs remains unexplored.

Enabling LLMs to reason over user preferences requires more than their general language
capabilities. Effective preference reasoning requires domain-specific knowledge
about item attributes, relationships, and up-to-date information that LLMs
inherently lack. Knowledge graphs (KGs) provide this essential structured knowledge
through rich semantic relationship and have proven effective in CRSs~\cite{chen_towards_2019,
zhou_improving_2020, wang_towards_2022, qiu_kerl_2025}. Despite their
demonstrated potential, integrating KGs with LLMs for preference reasoning presents
fundamental challenges that existing approaches have not addressed:
\begin{itemize}
  \item \textbf{Modality Gap (Challenge 1):} There exists a significant modality
    gap between KGs and LLMs. While LLMs process sequences of tokens representing
    natural language, KGs represent information in structured, graph-based formats
    with explicit entity relationships. This representational mismatch limits
    LLMs's ability to directly interpret and reason with the domain knowledge
    encoded in KGs.

  \item \textbf{Cross Modal Reasoning (Challenge 2):} Effectively reasoning over
    both the KG and conversation to infer user preferences is a complex task. LLMs,
    despite their strong natural language processing capabilities, are not inherently
    designed to perform this cross-modal reasoning. Therefore, they face difficulties
    in analyzing and synthesizing insights from graph-structured knowledge alongside
    dialogue history. This limitation hinders their ability to identify relevant
    patterns across both sources and to perform the knowledge-aware reasoning
    necessary for comprehensive user preference modeling.
\end{itemize}

To address the above challenges, we propose \textbf{Com}pact \textbf{P}reference
\textbf{A}nalyzer and \textbf{S}ummarization \textbf{S}ystem (COMPASS), a novel framework
that augments the LLM with the KG to enable explainable preference reasoning,
improving both recommendation performance and the explainability of existing
CRSs. COMPASS introduces a two-stage training paradigm that fundamentally transforms
how LLMs process and reason with structured knowledge for preference analysis.
First, we introduce \textit{graph entity captioning pre-training} that
transforms KG structures into natural language descriptions. This allows the LLM
to comprehend domain-specific information and bridge the \textbf{modality gap (Challenge
1)}. We employ a Graph Neural Network (GNN) to encode structural information from
the KG as entity embeddings, which are then projected into the LLM's semantic
space via an adapter module. The LLM learns to decode these projected embeddings
into textual descriptions of entities and their relationships, effectively
translating graph-structured knowledge into a format compatible with its
reasoning mechanisms. Building upon this alignment, COMPASS employs \textit{knowledge-aware
instruction fine-tuning} to enhance the LLM's ability to reason about user preferences
from dialogue histories and KG-augmented contexts. These KG-augmented contexts
consist of relevant entity information and relationships extracted from the KG, providing
a rich background for inference beyond the conversation history alone. Through
carefully designed instructions, we enhance the LLM's capability to perform \textbf{cross-modal
reasoning (Challenge 2)} by analyzing conversation history and cross-referencing
with KG-augmented information. This instruction tuning process enhances the LLM's
ability to extract explicit mentions, infer implicit interests, and reason about
preferences in relation to various item attributes. Consequently, as shown in the
bottom of Figure~\ref{fig:intro_fi}, COMPASS generates comprehensive and
interpretable user preferences in text that capture both overall preferences and
current interests. To leverage these insights, we introduce an adaptive gating
mechanism that integrates summarized preferences into existing CRSs, boosting recommendation
performance and explainability without requiring architectural changes. Our main
contributions can be summarized as follows:
\begin{itemize}
  \item \textbf{New framework.} We propose COMPASS, a novel framework for enhancing
    user preference modeling in CRSs. To the best of our knowledge, this is the first
    work to leverage LLMs and KGs for explainable preference generation in CRSs.

  \item \textbf{Effective cross-model reasoning and explanation.} We develop a
    two-stage process that enables the LLM to perform cross-modal reasoning over
    KGs and conversations, generating explainable user preference summaries.
    This approach moves beyond abstract vector representations to provide clear,
    human-readable user preferences.

  \item \textbf{Flexible plug-in.} COMPASS generates user preference summaries
    that are compatible with existing CRS architectures without requiring
    modifications to the system, improving both recommendation performance and explainability.
\end{itemize}
    \section{Related Work}
Knowledge graphs enable systems to infer implicit user interests and capture item relationships~\cite{chen_towards_2019, zhou_improving_2020},
% \subsection{Conversational Recommender System}
\noindent
\textbf{Conversational Recommender System.} Conversational recommender systems (CRSs) understand user preferences through interactive dialogue to provide personalized recommendations~\cite{jannach_survey_2021}.
These systems face a unique challenge in modeling user preferences from
typically brief and sparse conversational data. To address this, recent works
incorporate external knowledge sources into CRSs to enhance preference modeling.
Knowledge graphs enable systems to infer implicit user interests and capture item relationships~\cite{chen_towards_2019, zhou_improving_2020},,
while user reviews provide additional insights into item attributes and user
opinions~\cite{lu_revcore_2021, zhou_c-crs_2022}, helping to address
the challenge of limited contextual information. Moreover, pre-trained language models like DialoGPT~\cite{wang_towards_2022, wang-etal-2022-recindial} and BART~\cite{wang2022barcorunifiedframeworkconversational,qiu_kerl_2025} has further
improved contextual understanding and language generation capabilities in CRSs.

More recently, the advent of powerful LLMs has introduced new opportunities for
CRSs. Studies have focused on improving LLM evaluation through enhanced user simulators~\cite{wang-etal-2023-rethinking-evaluation,
    zhu_how_reliable_2024}, exploring zero-shot recommendation capabilities~\cite{he_large_language_2023},
    using graph-based retrieval augmentation for in-context learning~\cite{qiu_gcrs_2025},
and utilizing LLMs for task planning~\cite{feng_largelanguagemodelenhanced_2023,
    fang_multiagentconversa_2024}. However, one critical aspect that remains underexplored is the role of explainable user preference
in CRSs. Although some conceptual designs propose using LLMs to manage user
preferences~\cite{friedman_leveraginglargelanguagemodels_2023}, they lack quantitative validation. MemoCRS~\cite{xi_memocrs_2024} addresses user preference memories in
sequential CRS but does not fully harness LLM's reasoning abilities for
preference generation. Our work introduces the first framework that augments LLMs with KG for reasoning over user preferences in CRSs, enhancing both preference modeling and recommendation explainability.
Our work introduces the first framework that augments LLMs with KG for reasoning over user preferences in CRSs, enhancing both preference modeling and recommendation explainability.

\noindent
\textbf{Explainable Recommendation.} Explainable recommendation aims to enhance user trust and decision-making by providing transparent rationales for recommendations. Early research leveraged user-item interactions and content features to generate simple explanations~\cite{zhang_explainable_rec_2020},
using techniques like recurrent neural networks~\cite{donker_sequential_2017}, attention
mechanisms~\cite{chen_neural_att_2018}, and graph neural networks~\cite{wang_kgat_2019}.
Recent developments in LLMs have opened new avenues for explainable recommendation.
Recent LLMs have shown potential in generating explanations by analyzing user interactions~\cite{ma2024xrec, wang_can_2024}, but primarily operate within static user profiles where preferences are inferred from historical behavior. 
Our work introduces a novel KG-augmented LLM
framework for CRSs, enhancing both preference modeling and recommendation
explainability in dynamic interaction contexts.

    \section{Preliminaries}
\begin{figure*}[ht]
  \centering
  \includegraphics[width=\linewidth]{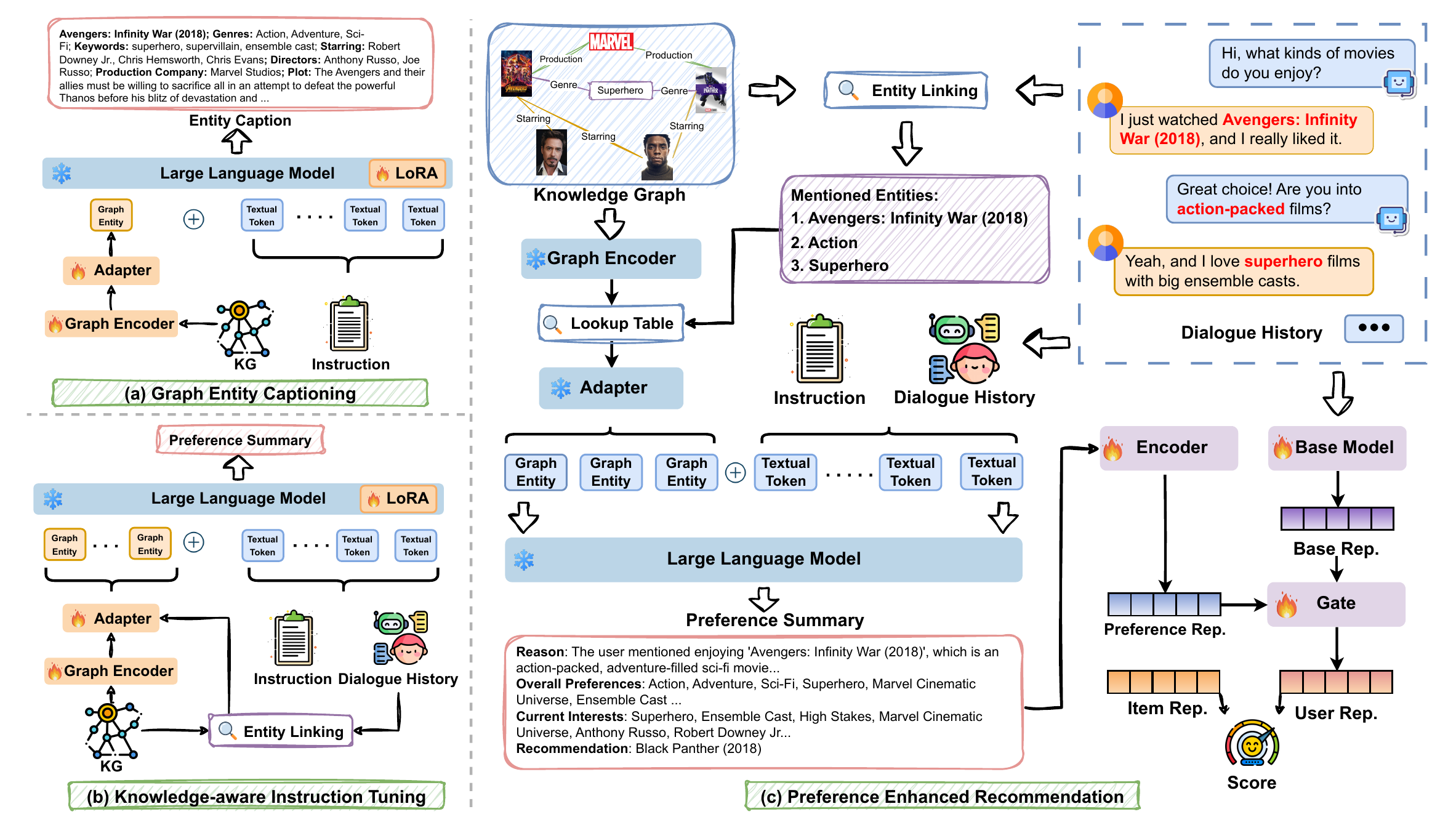}
  \vspace{-0.65cm}
  \caption{The overall framework of our COMPASS. COMPASS consists of three components: the graph encoder, the adapter, and the LLM. The adapter aligns the knowledge graph to the LLM. COMPASS follows a two-stage training paradigm - (a) \textit{Graph Entity Captioning} and (b) \textit{Knowledge-aware Instruction Tuning}. Once trained, COMPASS can be easily integrated with existing CRS models for \textit{(c) Preference Enhanced Recommendation}.}
  \label{fig:compass}
  \vspace{-0.4cm}
\end{figure*}
\noindent\textbf{Conversational Recommendation}.
Let $i$ denote a candidate item from the set of items $\mathcal{I}$, and let $w$ denote a word in the vocabulary $\mathcal{V}$.
A dialogue $\mathcal{D}$ between a user and the recommender system consists of a sequence of utterances $\mathcal{D} = [u_{1}, \ldots, u_{T}]$, where $u_t = [w_{1}^{t}, \ldots, w_{m}^{t}]$
is the $t$-th utterance composed of $m$ words, and $T$ is the maximum number of turns in the dialogue.
As the conversation progresses, the dialogue history up to the turn $t$ is denoted as $H_t = [u_1:u_t]$, where $[u_1:u_t]$ signifies the chronological sequence of utterances from the first to the $t$-th turn.
The CRS estimates the user's preferences based on $H_t$, then recommend $K$ items from $\mathcal{I}$, which are used to generate the next utterance $u_{t+1}$. Note that $\mathcal{I}_t$ can be empty when no recommendation is needed. In such cases, the CRS may raise a clarification question or generate a casual conversation response.

\noindent\textbf{Knowledge Graph}.
A knowledge graph is defined as $\mathcal{G} = (\mathcal{E}, \mathcal{A}, \mathcal{X})$ where $\mathcal{E},\mathcal{R}$ represents the set of entities and relation types in the graph, respectively. $\mathcal{A}$ is the adjacency matrix capturing the relationships between entities and $\mathcal{X}$ represents the textual descriptions of each entity.
For each entity $e \in \mathcal{E}$, its description is denoted as $x_e \in \mathcal{X}$, where $x_e = [w_{1}^{e}, \ldots, w_{k}^{e}]$, and $w_{k}^{e}$ represents the $k$-th word in the entity description.
% Each entity $e \in\mathcal{E}$ is associated with a textual description $x_e = \{w_{1}^{e}, \ldots, w_{k}^{e}\}$, where $w_{k}^{e}$ represents the $k$-th word in the description of entity $e$.
The entity set $\mathcal{E}$ encompasses candidate items $\mathcal{I}$ (e.g., movies) and non-item entities that represent item attributes (e.g., actors, genres, keywords). Formally, $\mathcal{I} \subseteq \mathcal{E}$.

\noindent\textbf{Explainable User Preferences and Recommendations}.
Explainable user preferences are crucial for enhancing the transparency and effectiveness of CRSs.
Our goal is to generate clear, human-understandable textual user preference summaries that provide insights for both recommendation and explanation.
Specifically, for a given dialogue history $H_t$, we define the generation of user preference summaries as:
\begin{equation}
  \setlength\abovedisplayskip{1pt}
  \setlength\belowdisplayskip{1pt}
  \mathcal{P}_{t} = f(I_p, H_t, \mathcal{E}_{t}^{m}, \mathcal{G}),
\end{equation}
where $f$ represents a model that reasons over the dialogue history $H_t$, the mentioned entities $\mathcal{E}_{t}^{m} \subseteq \mathcal{E}$, and their associated information from the knowledge graph $\mathcal{G}$. $I_p$ is denoted as instruction prompts. The resulting $\mathcal{P}_t$ represents the textual explainable user preference summary at the $t$-th turn.
The summary $\mathcal{P}_t$ is then encoded using a preference encoder $g(\cdot)$ and integrated into the base CRS model $f_{\text{crs}}$, which uses it to adjust its recommendation strategy.
Depending on the model, additional inputs, such as KG information or dialogue history, may also be used. Formally, the recommendation step is represented as:
\begin{equation}
  \setlength\abovedisplayskip{1pt}
  \setlength\belowdisplayskip{1pt}
  \mathcal{I}_t = f_{\text{crs}}(g(\mathcal{P}_t), H_t, \mathcal{G}),
\end{equation}
where $\mathcal{I}_t$ represents the recommended items at turn $t$, and $g(\mathcal{P}_t)$ denotes the encoded preference summary.
    \section{Methodology}

\subsection{Overview}
The primary goal of COMPASS is to synergize the reasoning capabilities of LLMs with
the structured knowledge from KGs to analyze and summarize user preferences.
COMPASS comprises three core components: (1) \textit{Graph encoder} processes a
domain-specific KG, capturing complex relationships between items to augment user
preference modeling. (2) \textit{Graph-to-Text adapter} aims to bridge the modality
gap between the graph encoder and the LLM, enabling the LLM to comprehend the
graph structure and conduct reasoning. (3) \textit{Large Language Model (LLM)}
leverages the powerful reasoning and generative capabilities of advanced
language models to generate interpretable user preference summaries.

To integrate the components cohesively, we employ a two-stage training process:
(1) \textit{Graph entity captioning} aligns KG structures with natural language representations,
creating a shared semantic space for the LLM to comprehend and reason with
domain-specific knowledge effectively. (2) \textit{Knowledge-aware instruction
fine-tuning} optimizes the LLM for cross-modal reasoning, allowing it to generate
comprehensive user preference summaries by synthesizing information from dialogue
history and KG-augmented context. Once trained, COMPASS can be integrated with existing
CRS models to improve their recommendation performance and explainability by generating
user preferences and incorporating them into the recommendation process with an adaptive
gating mechanism. Figure~\ref{fig:compass} illustrates COMPASS's architecture
and training process.

\subsection{Model Architecture}

\subsubsection{Graph Encoder}
In COMPASS, the KG is a crucial source of domain-specific information that
provides extra context for understanding attributes and relationships of items. To
efficiently convert the structured knowledge in the KG into a format understandable
by the LLM for preference reasoning and summarization, we utilize a Relational
Graph Convolutional Network (R-GCN)~\cite{schichtkrull_modeling_2018} to capture
the complex graph structure and generate entity embeddings. The R-GCN is well-suited
for modeling KGs due to its ability to handle multi-relational data and capture higher-order
dependencies between entities.

To initialize the entity embeddings, we leverage the textual descriptions
$\mathcal{X}$ associated with the entities in the KG.
Specifically, for each entity $e \in \mathcal{E}$, we encode its description $x_{e}$
using a PLM. This provides a rich semantic foundation for the graph learning
process. The R-GCN then captures both entity-level information and the overall graph
structure through iterative message passing, which is particularly important for
understanding the relationships between items and attributes. Formally, the
representation of an entity $e$ at the $l$-th layer is calculated as follows:
\begin{equation}
    \mathbf{h}_{e}^{(0)}= \text{PLM}(x_{e}), 
    \label{eq:rgcn_init}
\end{equation}
\begin{equation}
    \mathbf{h}_{e}^{(l+1)} = \sigma \left( \sum_{r\in\mathcal{R}}\sum_{e'\in{\mathcal{E}_{e}^{r}}}\frac{1}{Z_{e,r}}\mathbf{W}_{r}^{(l)}\mathbf{h}_{e'}^{(l)} + \mathbf{W}_{e}^{(l)}\mathbf{h}_{e}^{(l)} \right),
    \label{eq:rgcn}
\end{equation}

where $\mathbf{h}_{e}^{l}$ is the embedding of entity $e$ at the $l$-th layer,
${\mathcal{E}_{e}^{r}}$ is the set of neighboring entities connected to $e$
through relation $r$, $\mathbf{W}_{r}^{(l)}$ and $\mathbf{W}_{e}^{(l)}$ are learnable
weight matrices, ${Z_{e,r}}$ is a normalization factor, and $\sigma$ is an activation
function.

\subsubsection{Graph-to-Text Adapter}
\label{sec:graph-to-text} The entity embeddings from the graph encoder exist in a different representational space from the LLM's textual token representations. This makes it challenging for the
LLM to effectively reason with the KG-augmented context to generate user
preferences. To bridge this semantic gap and enable effective knowledge integration,
we introduce an adapter module that creates a mapping between graph-structured
entity embeddings and the LLM's textual domain. Specifically, the adaptation
process is defined as:
\begin{equation}
    \mathbf{h}_{e}^{\tau}= f_{P}(\mathbf{h}_{e}),\label{eq:proj}
\end{equation}
where $\mathbf{h}_{e}^{\tau}$ is the adapted entity embedding aligned with the LLM's
semantic space, $\mathbf{h}_{e}$ is the entity embedding generated from graph
encoder, and $f_{P}$ is the projection function implemented as a linear layer.
\subsubsection{Large Language Model (LLM)}

The LLM is the main reasoning engine that generates user preferences by synthesizing
information from dialogue histories and KG-augmented contexts. It captures both
the explicit user preferences expressed in the dialogue and the implicit
preferences inferred from the KG-augmented context for accurate preference
reasoning. Our framework is compatible with various state-of-the-art LLMs,
allowing flexibility in model choice. 
% In this paper, we use Llama3.1-8B \cite{grattafiori2024llama3herdmodels}
% for its natural language understanding and generation capabilities.

\subsection{Training Pipeline}
\subsubsection{Graph Entity Captioning}
To obtain a Graph-to-Text adapter that bridges the modality gap
between the graph entity embeddings and the LLM's semantic space, inspired by the
pre-training strategies employed in vision-language models~\cite{dai_instruct_blip_2023,
liu_visual_ins_2023}, we introduce a graph entity captioning pre-training
mechanism, as shown in Figure~\ref{fig:compass}a. This process creates a strong connection
between graph-structured data and natural language by generating entity-specific
captions that combine intrinsic entity information with neighboring node data, simulating message-passing within the graph structure. The pre-training stage enables the LLM to interpret graph entity
embeddings in a semantic context, facilitating deeper understanding of KG relationships and attributes for preference modeling.

The caption generation process differentiates between item entities (e.g.,
movies) and non-item entities (e.g., genres, directors). For item entities, we
employ a structured template that captures key attributes and relationships.
\vspace{-0.1cm}
\begin{tcolorbox}
    [colback=yellow!10!white, colframe=yellow!60!black, fontupper=\small, fonttitle=\small,
    boxsep=2pt, title=Example of Item Entity Caption Template] \textit{\textbf{Movie
    Title:} \textless Title\textgreater; \textbf{Genres:} \textless Genres\textgreater;
    \textbf{Keywords:} \textless Keywords\textgreater; \textbf{Starring:}
    \textless Actors\textgreater; \textbf{Directors:} \textless Directors\textgreater;
    \textbf{Production Company:} \textless Company\textgreater; \textbf{Plot:} \textless
    Plot Summary\textgreater}
\end{tcolorbox}
\vspace{-0.1cm}
\noindent
This template ensures comprehensive coverage of item attributes while maintaining
a consistent structure across different items. For non-item entities, we adopt a
more flexible approach that emphasizes the entity's role and its connections within
the KG.

To format the training data, an input-output pair is constructed for each entity
in the knowledge graph. The input consists of the adapted entity embedding $\mathbf{h}
_{e}^{\tau}$ from Equation~\ref{eq:proj} and a task-specific instruction prompt $I
_{c}$. The output is the generated caption $C_{e}$ for the entity $e$. This format
is represented as follows.
\vspace{-0.1cm}
\begin{tcolorbox}
    [colback=cyan!10!white, colframe=cyan!60!black, fontupper=\small, fonttitle=\small,
    title=Entity Captioning] \label{box:entity_caption} \textbf{Input:}
    \textless Entity Embeddings $\mathbf{h}_{e}^{\tau}$ \textgreater, \textless Instruction
    $I_{c}$\textgreater \\ \textbf{Output:} \textless Entity Caption $C_{e}$\textgreater
\end{tcolorbox}
\vspace{-0.1cm}

\noindent
In this way, the LLM learns to map the graph-structured input to natural
language by reconstructing the entity caption, conditioned on the graph entity embeddings
and instructions. This process is optimized by minimizing the negative log-likelihood
(NLL) of the generated captions, as expressed by:
\begin{equation}
    \mathcal{L}_{caption}= - \sum_{e \in \mathcal{E}}\log P(C_{e}\mid \mathbf{h}_{e}
    ^{\tau}, I_{c}),
\end{equation}
where $\mathcal{E}$ is the set of all entities in the KG. Through the training,
the LLM learns to interpret graph entity embeddings in a semantic context, leading
to a deeper understanding of the relationships and attributes encoded within the
KG. This understanding is essential for improving subsequent preference modeling.
\subsubsection{Knowledge-aware Instruction Tuning}

After the graph entity captioning pre-training, the LLM has gained a basic understanding
of the KG structure and content. However, it has not yet been explicitly trained
to utilize this knowledge for downstream tasks such as preference modeling and recommendation
generation. To this end, we introduce knowledge-aware instruction tuning to enable the LLM to reason across modalities, integrating dialogue history and KG information to infer user preferences and interests, as shown in Figure~\ref{fig:compass}b.

The knowledge-aware tuning process employs a carefully crafted instruction prompt~$I
_{p}$ to guide the LLM in synthesizing and reasoning over inputs from multiple sources.
Given a dialogue history~$H_{t}$ and the entities~$\mathcal{E}_{t}^{m}$
mentioned within it, the process retrieves the embeddings~$\mathbf{E}_{t}$ of these
entities from Equation~\ref{eq:proj}. These embeddings, along with the full dialogue
history, serve as inputs for the LLM. The instruction prompt~$I_{p}$ directs the
LLM to analyze both the KG-derived entity information and the dialogue history to
generate a user preference summary~$\mathcal{P}_{t}$.

This prompt follows a coarse-to-fine structure containing four key steps: (1)
reasoning, providing transparency in the model's decision-making; (2) overall
preferences, offering a broad view of the user's tastes; (3) current interests, capturing
recent and specific preferences to guide subsequent recommendations; and (4)
recommendation, leveraging the LLM's reasoning capabilities to suggest relevant items
aligned with user preferences, guiding downstream CRS models. This structured prompt
ensures that the model captures both long-term preferences and immediate interests.
To generate ground-truth preference summaries, we utilize an advanced LLM (e.g.,
ChatGPT) that performs cross-modal reasoning, integrating complete dialogue
histories with structured metadata of mentioned items from the KG.
% The detailed process and instruction templates for this ground-truth generation are provided in Appendix \ref{app:ground-truth}.
The instruction tuning process can be represented in the following format:
\vspace{-0.1cm}
\begin{tcolorbox}
    [colback=cyan!10!white, colframe=cyan!60!black, fontupper=\small, fonttitle=\small,
    boxsep=2pt, title=Knowledge-aware Instruction Tuning] \textbf{Input:} \textless
    Mentioned Entities Embeddings $\mathbf{E}_{t}$\textgreater, \textless Instruction
    $I_{p}$\textgreater, \textless Dialogue History $H_{t}$\textgreater \\
    \textbf{Output:} \textless Preference Summary $\mathcal{P}_{t}$\textgreater
\end{tcolorbox}
\vspace{-0.1cm}
% \noindent A detailed example of the preference summary can be found in Table~\ref{tab:case_study} of Appendix~\ref{app:case_study}.
\noindent
A detailed example of the preference summary can be found in Table~\ref{tab:case_study}.
\noindent
This instruction tuning process is optimized by minimizing the NLL of the generated
preference summaries, as expressed by:
\begin{equation}
    \mathcal{L}_{preference}= -\sum_{\mathcal{D}\in\mathcal{C}}\sum_{H_t\in\mathcal{D}}
    \log P(\mathcal{P}_{t}\mid \mathbf{E}_{t}, I_{p}, H_{t}),
\end{equation}
where $\mathcal{C}$ represents the set of all dialogues in the training data.
Through this process, COMPASS learns to synthesize information from dialogue history
and KG-derived entity embeddings, enabling it to generate comprehensive and
interpretable preference summaries.

\subsection{Integration with Existing CRS Models}
\label{sec:preference_enhanced_recommendation} To enhance the existing CRS
models with the user preference summaries generated by COMPASS, we propose a two-step
integration process: (1) transforming the natural language preference summaries
into a format compatible with CRS models, and (2) incorporating these transformed
preferences to enhance the base CRS model's recommendation performance via an adaptive
gating mechanism. Note that COMPASS remains frozen during this process.

\subsubsection{Preference Representation}
Traditional CRSs are not designed to directly utilize natural language
preference summaries. We employ a PLM-based encoder to transform these summaries
into a format suitable for existing CRS architectures. The encoder is adaptable
and can be implemented as either a frozen or trainable text encoding model. We employ
BERT~\cite{devlin_bert_2019} to encode the user preference summary as follows:

\begin{equation}
    \bm{s}_{c}= \text{PLM}(\mathcal{P}),
\end{equation}
where $\bm{s}_{c}$ is the encoded preference from text preference $\mathcal{P}$,
specifically the [CLS] token embedding. The contextual understanding of language
models enables $\bm{s}_{c}$ to capture comprehensive user preference summaries at
both coarse-grained (i.e., overall user preferences) and fine-grained levels (i.e.,
current interests and specific items).

\subsubsection{Enhanced Recommendation}

We integrate the encoded preferences into existing CRS models to enhance their
recommendation performance. We employ an adaptive gating mechanism to enhance the
preference representation $\bm{s}_{b}$ captured from the base CRS model
\footnote{The
    preference representation $\bm{s}_{b}$ is specific to each CRS model as
    different models employ distinct methods for building user preferences.}
with our COMPASS-generated representation $\bm{s}_{c}$. Note that the preference
representation $\bm{s}_{b}$ is specific to each CRS model as different models
employ distinct methods for building user preferences. Formally, we have: {
\begin{align}\label{eq:gating}\gamma&= \sigma(\mathbf{W}[\bm{s}_{b}; \bm{s}_{c}]), \\ \bm{s}_{u}&= \gamma\bm{s}_{b}+ (1 - \gamma)\bm{s}_{c},\end{align} where $\mathbf{W}$ are the learnable weight matrices, $\sigma$ is the sigmoid activation function, and ${\gamma}$ represents the gating probability. This adaptive mechanism controls the influence of each representation on the final user preference representation $\bm{s}_{u}$. The recommendation score for each item is computed using dot-product similarity as: }
\begin{equation}
    \label{eq:rec}%
    \setlength{\abovedisplayskip}{2pt}
    \setlength{\belowdisplayskip}{2pt}
    P_{rec}(i) = \text{softmax}(\bm{s}_{u}\cdot\bm{I}_{i}),
\end{equation}
where $\bm{I}_{i}$ are the item representations\footnote{The item
representations $\bm{I}_{i}$ are specific to each CRS model and may vary
depending on the architecture employed by the base CRS.}. Finally, we optimize
the recommendation loss $\mathcal{L}_{\text{rec}}$ as follows:

\begin{equation}
    \mathcal{L}_{\text{rec}}= - \sum_{j=1}^{N}\sum_{i=1}^{M}y_{ij}\log \left( \bm
    {P}_{\text{rec}}^{(j)}(i) \right),\label{eq:rec_loss}
\end{equation}
where $N$ is the number of conversations, $M$ is the number of items, and
$y_{ij}$ is the ground truth label indicating whether item $i$ is relevant to
conversation $j$.
    \section{Experiments}

\subsection{Experimental Settings}

\subsubsection{Datasets}
\begin{table*}
    [t!]
    \centering
    % \caption{Performance comparison on recommendation tasks. `COM.' denotes
    % models enhanced with the COMPASS approach and `Improv.' indicates the relative
    % improvement of the COMPASS compared to the original base model. The best
    % results are highlighted in bold ($\mathbf{t}$-test with $\mathbf{p}$-value $<
    % \mathbf{0.05}$).}
    \caption{Performance comparison on recommendation tasks. `COM.' denotes
    models enhanced with the COMPASS approach and `Improv.' indicates the relative
    improvement of COMPASS compared to the original base model. The best results
    are highlighted in bold (t-test with $p < 0.05$).}
    \vspace{-0.1cm}
    \label{tab:rec_results}
    \resizebox{1\linewidth}{!}{%
    \begin{tabular}{@{}llcccccccccccc@{}}
        \toprule \multirow{2}{*}{Types}        & \multirow{2}{*}{Model}                                     & \multicolumn{6}{c}{ReDial} & \multicolumn{6}{c}{INSPIRED} \\
        \cmidrule(lr){3-8} \cmidrule(lr){9-14} &                                                            & HR@10                      & HR@50                       & NDCG@10        & NDCG@50        & MRR@10         & MRR@50         & HR@10          & HR@50          & NDCG@10        & NDCG@50        & MRR@10         & MRR@50         \\
        \midrule \multirow{11}{*}{Base}        & ReDial~\cite{li_towards_2018}                              & 0.140                      & 0.320                       & 0.061          & 0.065          & 0.035          & 0.045          & 0.117          & 0.285          & 0.035          & 0.072          & 0.022          & 0.048          \\
                                               & BERT~\cite{devlin_bert_2019}                               & 0.143                      & 0.319                       & 0.073          & 0.108          & 0.052          & 0.059          & 0.179          & 0.328          & 0.095          & 0.125          & 0.072          & 0.079          \\
                                               & GPT-2~\cite{radford2019language}                           & 0.147                      & 0.327                       & 0.071          & 0.107          & 0.051          & 0.056          & 0.112          & 0.278          & 0.089          & 0.128          & 0.063          & 0.076          \\
                                               & Llama3.1-8B~\cite{grattafiori2024llama3herdmodels}         & 0.188                      & 0.376                       & 0.103          & 0.146          & 0.078          & 0.087          & 0.190          & 0.332          & 0.118          & 0.150          & 0.094          & 0.102          \\
                                               & KBRD~\cite{chen_towards_2019}                              & 0.151                      & 0.336                       & 0.099          & 0.136          & 0.071          & 0.079          & 0.172          & 0.265          & 0.106          & 0.127          & 0.086          & 0.091          \\
                                               & KGSF~\cite{zhou_improving_2020}                            & 0.183                      & 0.378                       & 0.098          & 0.140          & 0.072          & 0.081          & 0.175          & 0.273          & 0.106          & 0.128          & 0.088          & 0.093          \\
                                               & PECRS~\cite{ravaut-etal-2024-parameter}                    & 0.205                      & 0.399                       & 0.112          & 0.154          & 0.083          & 0.093          & 0.179          & 0.337          & 0.106          & 0.142          & 0.084          & 0.092          \\
                                               & BARCOR~\cite{wang2022barcorunifiedframeworkconversational} & 0.169                      & 0.374                       & 0.088          & 0.133          & 0.063          & 0.073          & 0.185          & 0.339          & 0.104          & 0.137          & 0.080          & 0.087          \\
                                               & UniCRS~\cite{wang_towards_2022}                            & 0.216                      & 0.416                       & 0.118          & 0.163          & 0.087          & 0.095          & 0.250          & 0.408          & 0.182          & 0.218          & 0.148          & 0.160          \\
                                               & DCRS~\cite{dao_dcrs_2024}                                  & 0.242                      & 0.422                       & 0.142          & 0.182          & 0.111          & 0.120          & 0.225          & 0.417          & 0.152          & 0.194          & 0.128          & 0.135          \\
                                               & MSCRS~\cite{wei_mscrs_2025}                                & 0.255                      & 0.448                       & 0.153          & 0.192          & 0.122          & 0.130          & 0.257          & 0.424          & 0.163          & 0.205          & 0.145          & 0.151          \\
        \midrule \multirow{10}{*}{Enhanced}    & COM.+KBRD                                                  & 0.199                      & 0.412                       & 0.103          & 0.150          & 0.075          & 0.085          & 0.249          & 0.392          & 0.152          & 0.183          & 0.123          & 0.129          \\
                                               & \quad +Improv.                                             & 31.79\%                    & 22.61\%                     & 4.04\%         & 10.29\%        & 5.63\%         & 7.59\%         & 44.76\%        & 47.92\%        & 43.39\%        & 44.09\%        & 43.02\%        & 38.70\%        \\
        \cmidrule(lr){2-14}                    & COM.+KGSF                                                  & 0.198                      & 0.413                       & 0.105          & 0.152          & 0.076          & 0.088          & 0.197          & 0.400          & 0.125          & 0.167          & 0.103          & 0.110          \\
                                               & \quad+Improv.                                              & 8.20\%                     & 9.26\%                      & 7.14\%         & 8.57\%         & 5.5\%          & 8.64\%         & 12.57\%        & 46.52\%        & 17.92\%        & 30.47\%        & 17.05\%        & 18.28\%        \\
        \cmidrule(lr){2-14}                    & COM.+BERT                                                  & 0.182                      & 0.382                       & 0.098          & 0.142          & 0.073          & 0.082          & 0.207          & 0.345          & 0.129          & 0.157          & 0.105          & 0.110          \\
                                               & \quad+Improv.                                              & 27.27\%                    & 19.75\%                     & 34.25\%        & 31.48\%        & 40.38\%        & 38.98\%        & 15.64\%        & 5.18\%         & 35.79\%        & 25.60\%        & 45.83\%        & 39.24\%        \\
        \cmidrule(lr){2-14}                    & COM.+Llama3.1-8B                                           & 0.215                      & 0.406                       & 0.118          & 0.161          & 0.089          & 0.100          & 0.232          & 0.377          & 0.146          & 0.169          & 0.117          & 0.122          \\
                                               & \quad +Improv.                                             & 13.76\%                    & 7.98\%                      & 14.45\%        & 14.10\%        & 12.82\%        & 14.94\%        & 22.11\%        & 13.55\%        & 23.73\%        & 12.67\%        & 24.47\%        & 19.61\%        \\
        \cmidrule(lr){2-14}                    & COM.+MSCRS                                                 & \textbf{0.264}             & \textbf{0.451}              & \textbf{0.160} & \textbf{0.200} & \textbf{0.129} & \textbf{0.137} & \textbf{0.277} & \textbf{0.427} & \textbf{0.183} & \textbf{0.214} & \textbf{0.154} & \textbf{0.160} \\
                                               & \quad +Improv.                                             & 3.53\%                     & 0.67\%                      & 4.58\%         & 4.17\%         & 5.74\%         & 5.38\%         & 7.78\%         & 0.71\%         & 12.27\%        & 4.39\%         & 6.21\%         & 5.96\%         \\
        \bottomrule
    \end{tabular}
    }
    \vspace{-0.1cm}
\end{table*}
We conduct our experiments on two widely used English CRS datasets: ReDial~\cite{li_towards_2018}
and INSPIRED~\cite{hayati_inspired_2020} datasets. ReDial dataset, focused on movie
recommendations, contains 11,348 conversations and is constructed through
crowdsourcing on Amazon Mechanical Turk (AMT). The INSPIRED dataset contains 999
conversations, also about movie recommendations, but additionally provides recommendation
strategies based on social science theories. All experiments
are conducted using the official dataset splits.
We constructed the KG
by scraping data from IMDB\footnote{https://www.imdb.com/}, using movie titles
and release years as key search terms.

\subsubsection{Baselines}
\label{sec:baselines} We consider a comprehensive range of baseline models, including
traditional CRS \textbf{ReDial}~\cite{li_towards_2018}, knowledge-graph based methods
such as \textbf{KBRD}~\cite{chen_towards_2019} and \textbf{KGSF}~\cite{zhou_improving_2020},
language model-based methods like \textbf{BERT}~\cite{devlin_bert_2019},
\textbf{GPT-2}~\cite{radford2019language}, and \textbf{Llama3.1-8B}~\cite{grattafiori2024llama3herdmodels},
metadata-augmented language model \textbf{PECRS}~\cite{ravaut-etal-2024-parameter},
and hybrid approaches that combine KGs with language models: \textbf{BARCOR}~\cite{wang2022barcorunifiedframeworkconversational}
fuses BART with a KG, \textbf{UniCRS}~\cite{wang_towards_2022} employs KG prompt
learning, \textbf{DCRS}~\cite{dao_dcrs_2024} leverages retrieval-augmented
prompt learning, and \textbf{MSCRS}~\cite{wei_mscrs_2025} introduces multi-modal
graph prompt learning. Since no existing baselines specifically generate user preference
summaries, we use \textbf{Llama3.1-8B} and \textbf{GPT-4o} as baselines for
comparison, both without KG access. For clarity, we denote \textbf{Llama3.1-8B}
as \textbf{Llama-Summary} when used for preference summary generation.

\subsubsection{Evaluation Metrics}
\label{sec:eval} Our evaluation framework assesses both the performance of the recommendations
and the quality of generated user preference summaries. For recommendation tasks,
we adopt widely used metrics from previous works~\cite{zhou_c-crs_2022, zhou_improving_2020},
including HR@$K$, NDCG@$K$, and MRR@$K$, with $K$ set to 10 and 50. To evaluate
user preference summaries, we employ three complementary metrics: (1) Lexical
Similarity: ROUGE-1, ROUGE-2, ROUGE-L, and ROUGE-Sum; (2) Semantic Understanding:
Following~\cite{wang-etal-2023-chatgpt}, we use GPT-4o-mini to assess Reasoning Proficiency
(RP) and Factual Consistency (FC). RP assesses logical coherence, inference accuracy,
and recommendation relevance, while FC evaluates dialogue alignment, knowledge
graph consistency, and statement accuracy. Both metrics are scored on a scale of
0 to 100. (3) User-Centric Evaluation: Inspired by~\cite{ji2023exploringimpactinstructiondata,
gao2024llmbasednlgevaluationcurrent}, we conduct LLM-based simulated user
evaluations, which rank the system's output based on two aspects: Explainability,
which measures how clearly the system justifies its recommendations, and User
Preference Alignment, which measures how accurately the system understands and responds
to user preferences expressed in the conversation. 

\subsubsection{Implementation Details}
We implement COMPASS using the Llama3.1-8B model~\cite{grattafiori2024llama3herdmodels}
as the base LLM, freeze all base LLM parameters and employ LoRA~\cite{hu2022lora}
for efficient fine-tuning. The graph encoder uses 1 layer with 768 hidden dimensions.
For entity linking, we use provided entities for INSPIRED dataset and GPT-3.5 to
extract entities from Redial dialogues, then apply text-embedding-ada-002 for
semantic matching to KG entities. We use batch sizes of 256 for pre-training and
128 for fine-tuning on Inspired dataset. For ReDial dataset, we maintain a batch
size of 128 for both pre-training and fine-tuning. All experiments run on Nvidia
A100 GPUs. For ground truth preference summaries, we employ GPT-4 to generate
summaries based on dialogue histories and associated item metadata from the KG
using structured instruction templates\footnote{Code, data, and instruction templates are available at: \url{https://github.com/icedpanda/COMPASS-official}}.
\subsection{Recommendation Evaluation}
\subsubsection{Improvement over Baseline Models}

COMPASS is designed to be flexible, allowing integration with different CRS models.
We evaluated its effectiveness across various models, with performance results shown in Table~\ref{tab:rec_results}. We have the following
observations: (1) Baseline models incorporating external KGs, such as KBRD and KGSF,
consistently outperform simpler language model-based approaches like BERT and
GPT-2. This highlights the importance of structured knowledge in capturing user preferences and item relationships.
settings. (2) Llama3.1-8B demonstrates strong performance, surpassing BARCOR and
performing competitively with PECRS on ReDial dataset, indicating that the advanced
language understanding and extensive world knowledge of LLMs can effectively capture
user preferences, resulting in more accurate recommendations. (3) Advanced hybrid
methods that integrate language models with KGs achieve the strongest baseline performance,
with MSCRS achieving the best results across all metrics on both datasets.

Building upon these baselines, COMPASS consistently enhances every integrated model regardless of architecture or performance level.
The framework delivers substantial gains on weaker models, achieving
44.76\% HR@10 improvement with KBRD on INSPIRED dataset and 27.27\% with BERT on
ReDial dataset. When integrated with LLMs like Llama3.1-8B, COMPASS maintains solid
enhancements of 13.76\% in HR@10 on ReDial dataset. 
Most importantly, COMPASS enhances the strongest baseline MSCRS, achieving 7.78\% HR@10 improvement and
reaching the best overall performance across all metrics. These consistent improvements across different architectures validate that COMPASS's knowledge-enriched preference representations provide universal value. These representations effectively
capture user intent while providing structured insights that enhance both
recommendation performance and explainability.

\subsubsection{Comparison of Enhancement Methods}
\begin{figure}[h]
    \centering
    \begin{subfigure}
        [b]{0.23\textwidth}
        \centering
        \includegraphics[width=\textwidth]{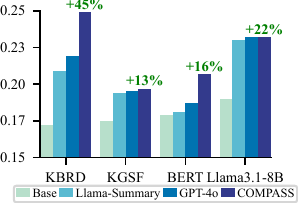}
        \caption{HR@10}
        \label{fig:hr@10}
    \end{subfigure}
    \begin{subfigure}
        [b]{0.23\textwidth}
        \centering
        \includegraphics[width=\textwidth]{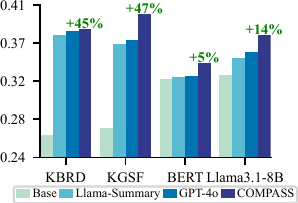}
        \caption{HR@50}
        \label{fig:hr@50}
    \end{subfigure}
    \hfill
    \begin{subfigure}
        [b]{0.23\textwidth}
        \centering
        \includegraphics[width=\textwidth]{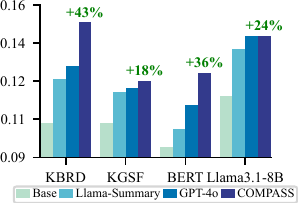}
        \caption{NDCG@10}
        \label{fig:ndcg@10}
    \end{subfigure}
    \begin{subfigure}
        [b]{0.23\textwidth}
        \centering
        \includegraphics[width=\textwidth]{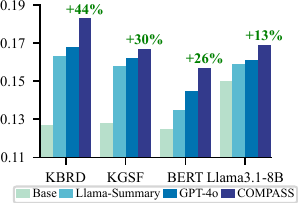}
        \caption{NDCG@50}
        \label{fig:ndcg@50}
    \end{subfigure}
    % \vspace{-0.2cm}
    \caption{Comparison of recommendation performance on the INSPIRED dataset
    with different enhancers. Green percentages show improvements over baselines.}
    \label{fig:enhancement_comparison}
    \vspace{-.2cm}
\end{figure}

To evaluate COMPASS's effectiveness, we compare it with Llama-Summary and GPT-4o
as baseline enhancers, which generate user preference summaries without KG
access. Figure~\ref{fig:enhancement_comparison} illustrates the performance of these
methods across various CRSs on INSPIRED dataset. While all enhancement methods
improve upon base models, COMPASS consistently outperforms both alternatives.
Notably, COMPASS surpasses GPT-4o despite the latter being a much larger model,
with improvements ranging from marginal gains on Llama3.1-8B to substantial improvements
on KBRD and KGSF. This superior performance demonstrates the critical importance
of integrating domain-specific knowledge into preference modeling. The substantial
performance gap between COMPASS and Llama-Summary highlights the effectiveness of
our framework in leveraging both structured knowledge and LLM capabilities.
These results show COMPASS's ability to generate more accurate and contextually
relevant user preference summaries, leading to improved recommendation performance
across different CRS architectures.

\subsubsection{Ablation study}

To assess the contribution of different components in COMPASS, we conducted an
ablation study by evaluating several model variants across three base models: KBRD,
KGSF, and LLaMA3.1-8B. We compare the COMPASS model against three ablated versions:
(1) \textbf{COM w/o REC}: The generated preference summary does not include
recommended items; (2) \textbf{COM w/o GEP}: COMPASS without the graph entity captioning
pre-training; (3) \textbf{COM w/o KG}: COMPASS without the graph encoder.

The results in Table \ref{tab:ab_rec_results} show that COMPASS generally
achieves the best performance across most metrics, highlighting its superior capability.
The COM w/o REC variant shows the largest performance drop, emphasizing the crucial
role of including recommended items in preference summaries to guide downstream models.
Both COM w/o KG and COM w/o GEP variants demonstrate comparable performance
declines, highlighting the crucial roles of KG integration for understanding item
relationships and GEP for comprehending KG-augmented context. These findings
show the importance of each COMPASS component, with their synergy driving the superior
performance of the full model.

\subsection{Preference Generation Evaluation}
To evaluate COMPASS's ability to reason over user preferences and generate explainable
summaries, we compare COMPASS with variants to assess key component effectiveness. Additionally, a case study with Llama-Summary and GPT-4o examples provides qualitative insights.
\newline
\subsubsection{Automatic Evaluation}

\begin{table}[t!]
    \centering
    \caption{Ablation study on recommendation task performance. The best-performing
    results are highlighted in bold.}
    \label{tab:ab_rec_results}
    % \vspace{-0.1cm}
    \resizebox{\linewidth}{!}{%
    \begin{tabular}{@{}lcccccc@{}}
        \toprule \multirow{2}{*}{Model} & \multicolumn{6}{c}{ReDial} \\
        \cmidrule(lr){2-7}              & HR@10                     & HR@50          & NDCG@10        & NDCG@50        & MRR@10         & MRR@50         \\
        \midrule KBRD                   & 0.151                     & 0.336          & 0.099          & 0.136          & 0.071          & 0.079          \\
        \quad +COM w/o KG               & 0.195                     & 0.405          & 0.102          & 0.147          & 0.074          & 0.084          \\
        \quad +COM w/o REC              & 0.190                     & 0.395          & 0.100          & 0.146          & 0.075          & 0.083          \\
        \quad +COM w/o GEP              & 0.188                     & 0.410          & 0.100          & 0.148          & 0.072          & 0.082          \\
        \quad +COMPASS                  & \textbf{0.199}            & \textbf{0.412} & \textbf{0.103} & \textbf{0.150} & \textbf{0.075} & \textbf{0.085} \\
        \midrule KGSF                   & 0.183                     & 0.378          & 0.098          & 0.140          & 0.072          & 0.081          \\
        \quad +COM w/o KG               & 0.196                     & 0.410          & 0.102          & 0.150          & 0.074          & 0.085          \\
        \quad +COM w/o REC              & 0.192                     & 0.408          & 0.101          & 0.150          & 0.074          & 0.084          \\
        \quad +COM w/o GEP              & 0.196                     & 0.406          & 0.102          & 0.149          & 0.073          & 0.084          \\
        \quad +COMPASS                  & \textbf{0.198}            & \textbf{0.413} & \textbf{0.105} & \textbf{0.152} & \textbf{0.076} & \textbf{0.088} \\
        \midrule Llama3.1-8B            & 0.188                     & 0.376          & 0.103          & 0.146          & 0.078          & 0.087          \\
        \quad +COM w/o KG               & 0.209                     & 0.399          & 0.116          & 0.159          & 0.087          & 0.096          \\
        \quad +COM w/o REC              & 0.202                     & 0.395          & 0.109          & 0.151          & 0.081          & 0.090          \\
        \quad +COM w/o GEP              & 0.212                     & 0.405          & 0.117          & \textbf{0.162} & 0.088          & 0.099          \\
        \quad +COMPASS                  & \textbf{0.215}            & \textbf{0.406} & \textbf{0.118} & 0.161          & \textbf{0.089} & \textbf{0.100} \\
        \bottomrule
    \end{tabular}
    }
    % \vspace{-0.1cm}
\end{table}
% \vspace{-0.2cm}
\begin{table}[t]
    \centering
    \caption{Evaluation of generated user preference summaries. RP is reasoning proficiency
    and FC is factual consistency. The best results are highlighted in bold.}
    \label{tab:preference_quality}
    % \vspace{-0.2cm}
    \resizebox{\linewidth}{!}{%
    \begin{tabular}{@{}lcccccc@{}}
        \toprule \multirow{2}{*}{Model} & \multicolumn{6}{c}{ReDial}   \\
        \cmidrule(lr){2-7}              & ROUGE-1                     & ROUGE-2        & ROUGE-L        & ROUGE-Sum      & RP             & FC             \\
        \midrule COM w/o KG             & 62.06                       & 39.16          & 49.91          & 59.33          & 81.33          & 82.13          \\
        COM w/o NP                      & 61.72                       & 38.95          & 49.74          & 59.00          & 81.16          & 82.54          \\
        COMPASS                         & \textbf{62.71}              & \textbf{40.61} & \textbf{51.14} & \textbf{60.03} & \textbf{82.20} & \textbf{84.21} \\
        \midrule \multirow{2}{*}{Model} & \multicolumn{6}{c}{INSPIRED} \\
        \cmidrule(lr){2-7}              & ROUGE-1                     & ROUGE-2        & ROUGE-L        & ROUGE-Sum      & RP             & FC             \\
        \midrule COM w/o KG             & 55.85                       & 32.94          & 43.36          & 53.12          & 82.59          & 84.33          \\
        COM w/o NP                      & 58.20                       & 34.57          & 44.73          & 55.08          & 82.36          & 84.29          \\
        COMPASS                         & \textbf{59.37}              & \textbf{35.93} & \textbf{46.88} & \textbf{56.64} & \textbf{83.14} & \textbf{85.17} \\
        \bottomrule
    \end{tabular}
    }
    % \vspace{-0.2cm}
\end{table}

\vspace{-0.3cm}
\begin{table*}
    [t!]
    \centering
    \caption{Case Study: Comparison of preference summaries generated by different
    models. Green indicates information supported by the KG, yellow indicates correct
    inferences not explicitly detailed in the KG but aligned with the content,
    and red indicates unsupported and unreasonable details.}
    \label{tab:case_study}
    \vspace{-0.1cm}
    \begin{tabular}{@{}p{0.02\textwidth}p{0.97\textwidth}@{}}
        \toprule \multirow{2}{*}{\rotatebox[origin=c]{90}{\textbf{Hist.}}}                                            & \textbf{System}: Hello! What kind of movies are you into?                                                                                                                                                                                                                                                                                                                                                                                                                                                                                                                                                                                                                                                                                                                                                                                                                                                                                                                                                                                                                                                                                                                                       \\
                                                                                                                      & \textbf{User}: I loved \textit{The Professional (1981)} and would love to see something similar.                                                                                                                                                                                                                                                                                                                                                                                                                                                                                                                                                                                                                                                                                                                                                                                                                                                                                                                                                                                                                                                                                                \\
        \midrule \multirow{4}{*}{\rotatebox[origin=c]{90}{\parbox{25em}{\centering\textbf{User Preference Summary}}}} & \textbf{COMPASS:} \textit{Reasoning:} The user mentioned `The Professional (1981)' as a movie they loved and would like to see something similar. By analyzing the knowledge graph, we can identify that `The Professional' falls under the genres of \sethlcolor{green}\hl{Action, Crime, and Thriller}, and includes themes such as \sethlcolor{yellow}\hl{professional killers, assassins, and intense action sequences}. The movie also features notable actors like \sethlcolor{green}\hl{Paul Belmondo and Michel Beaune}, and is directed by \sethlcolor{green}\hl{Georges Lautner}. These elements suggest the user enjoys \sethlcolor{yellow}\hl{high-stakes, action-packed narratives with complex characters and moral dilemmas.} \textit{Overall preferences}: \sethlcolor{green}\hl{Action, Crime, Thriller}, \sethlcolor{yellow}\hl{professional killers, assassins, intense action sequences, complex characters, moral dilemmas}, \sethlcolor{green}\hl{Jean-Paul Belmondo, Georges Lautner}. \textit{Current interests}: \sethlcolor{green}\hl{Action, Crime, Thriller}, \sethlcolor{yellow}\hl{professional killers, assassins, intense action sequences.}                    \\
        \cmidrule(l){2-2}                                                                                             & \textbf{GPT-4o:} \textit{Reasoning: }The user mentioned a specific movie, `The Professional (1981)', which they loved and expressed interest in finding similar movies. This indicates a preference for films with similar themes, styles, or elements. \textit{Overall preferences}: \sethlcolor{green}\hl{crime}, \sethlcolor{red}\hl{drama}, \sethlcolor{green}\hl{thriller}, \sethlcolor{red}\hl{intense character development}, \sethlcolor{yellow}\hl{complex relationships}, \sethlcolor{red}\hl{suspense}. \textit{Current interests}: \sethlcolor{green}\hl{The Professional (1981)}, \sethlcolor{red}\hl{crime drama}, \sethlcolor{green}\hl{thriller}, \sethlcolor{yellow}\hl{intense character dynamics.}                                                                                                                                                                                                                                                                                                                                                                                                                                                                           \\
        \cmidrule(l){2-2}                                                                                             & \textbf{Llama-Summary:} \textit{Reasoning}: The user's conversation history indicates a preference for \sethlcolor{green}\hl{1980s French cinema}, particularly from the \sethlcolor{red}\hl{Nouvelle Vague movement}, with an emphasis on \sethlcolor{yellow}\hl{intelligent}, stylish, and critically-acclaimed films that often involve \sethlcolor{yellow}\hl{morally complex situations.} \textit{Overall preferences}: \sethlcolor{red}\hl{French New Wave}, \sethlcolor{yellow}\hl{Intellectual}, Stylish, \sethlcolor{yellow}\hl{Morally Complex.} \textit{Current interests}: \sethlcolor{red}\hl{Jean-Pierre Mousel}, \sethlcolor{green}\hl{The Professional (1981)}, \sethlcolor{red}\hl{Claude Sautet, Claudie Claira.}                                                                                                                                                                                                                                                                                                                                                                                                                                                             \\
        \cmidrule(l){2-2}                                                                                             & \textbf{COM w/o GEP:} \textit{Reasoning}: The user mentioned `The Professional (1981)' as a favorite movie and expressed interest in finding something similar. By analyzing the knowledge graph, we can identify that `The Professional' falls under the genres of \sethlcolor{green}\hl{Action, Crime, and Thriller}, and includes themes such as \sethlcolor{red}\hl{police}, \sethlcolor{yellow}\hl{political corruption}, and \sethlcolor{red}\hl{organized crime}. The movie also features notable actors like \sethlcolor{red}\hl{Ryan O'Neal and Charles Grodin}, and is directed by \sethlcolor{red}\hl{Donald Shebib.} These elements suggest the user enjoys \sethlcolor{yellow}\hl{intense, plot-driven narratives with a focus on crime and political intrigue.} \textit{Overall preferences}: \sethlcolor{green}\hl{Action, Thriller}, \sethlcolor{red}\hl{police}, \sethlcolor{yellow}\hl{political corruption}, \sethlcolor{red}\hl{organized crime, Ryan O'Neal, Charles Grodin, Donald Shebib.} \textit{Current interests}: \sethlcolor{green}\hl{Thriller}, \sethlcolor{red}\hl{police}, \sethlcolor{yellow}\hl{political corruption}, \sethlcolor{red}\hl{organized crime.} \\
        \bottomrule
    \end{tabular}
    \vspace{-0.1cm}
\end{table*}

Table~\ref{tab:preference_quality} presents a comparative analysis of COMPASS
and its variants, COM w/o KG and COM w/o GEP, on ReDial and INSPIRED datasets. COMPASS
consistently outperforms its variants across all metrics on both datasets.
Lexical similarity metrics (ROUGE scores) demonstrate that COMPASS generates
summaries that align closely with reference texts, while higher reasoning proficiency
and factual consistency scores illustrate its superior reasoning quality and factual
accuracy. The performance gap between COMPASS and COM w/o KG demonstrates the
value of KG integration in enhancing reasoning over user preferences. Incorporating
structured domain knowledge enables more accurate preference inference and explainable
summary generation. Moreover, the comparative performance of COM w/o GEP and COMPASS
indicates that, despite the presence of structured knowledge from the KG, there remains
a modality gap between graph structures and natural language, posing challenges
for LLMs to effectively reason with KG information. These results validate the COMPASS
framework, demonstrating the synergistic effect of combining KGs with LLMs and
confirming the critical role of our graph entity captioning pre-training in
enabling effective cross-modal reasoning for explainable preference generation.

\subsubsection{LLM-Simulated User Evaluation}
\begin{figure}[t!]
    \centering
    \begin{subfigure}
        [b]{0.23\textwidth}
        \centering
        \includegraphics[width=\textwidth]{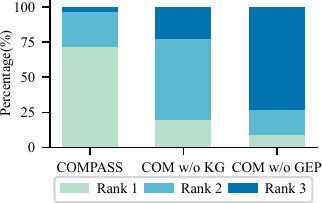}
        \caption{ReDial: Alignment}
        \label{fig:redial_alignment}
    \end{subfigure}
    \hfill
    \begin{subfigure}
        [b]{0.23\textwidth}
        \centering
        \includegraphics[width=\textwidth]{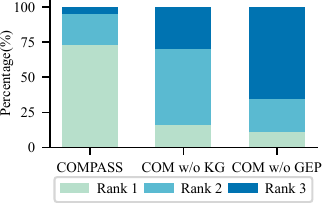}
        \caption{ReDial: Explainability}
        \label{fig:redial_explainability}
    \end{subfigure}
    \begin{subfigure}
        [b]{0.23\textwidth}
        \centering
        \includegraphics[width=\textwidth]{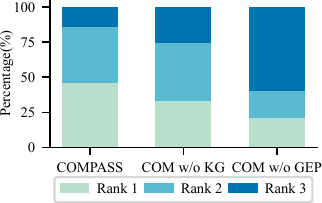}
        \caption{INSPIRED: Alignment}
        \label{fig:inspired_alignment}
    \end{subfigure}
    \hfill
    \begin{subfigure}
        [b]{0.23\textwidth}
        \centering
        \includegraphics[width=\textwidth]{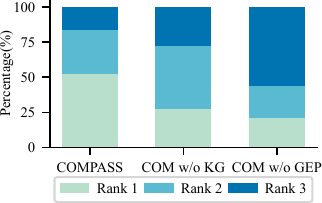}
        \caption{INSPIRED: Explainability}
        \label{fig:inspired_explainability}
    \end{subfigure}
    \vspace{-0.2cm}
    \caption{LLM-simulated user rankings for User Preference Alignment and
    Explainability across ReDial and INSPIRED datasets.}
    \label{fig:llm_eval}
    \vspace{-0.4cm}
\end{figure}

To assess user-perceived quality, we employ LLM-simulated user evaluations on explainability
and user preference alignment. Figure~\ref{fig:llm_eval} presents the results
across ReDial and INSPIRED datasets. COMPASS is consistently preferred over COM w/o
KG and COM w/o GEP in both metrics. On ReDial dataset, COMPASS is most preferred
in approximately 75\% of the cases for both alignment and explainability. Although
the preference gap is smaller on INSPIRED, COMPASS maintains a clear advantage. Notably,
COM w/o NP is consistently the least preferred, showing the importance of pre-training
in generating meaningful preference summaries. These results validate COMPASS's
effectiveness in reasoning over user preferences and generating user-friendly, explainable
summaries that enhance both fidelity and transparency in CRSs.

\subsubsection{Case Study}
We present a detailed case study in Table~\ref{tab:case_study} to demonstrate how
different models reason over user preferences and generate explainable summaries
from sample dialogue. COMPASS demonstrates compelling effectiveness in capturing
and reasoning about user preferences. It accurately identifies the movie genres and
lists key cast members and the director of \textit{The Professional (1981)}. Moreover,
COMPASS infers relevant themes that align with the movie's content, such as \textit{professional
killers}, and \textit{intense action sequences}. In contrast, other models show
varying levels of accuracy and reasoning depth. (1) GPT-4o, without KG
augmentation, provides a broader interpretation but less precise reasoning.
While it correctly identifies some genres, it includes unsupported elements,
illustrating that general LLM knowledge alone is insufficient for accurate
preference reasoning in domain-specific contexts. (2) Llama-Summary demonstrates
significant inaccuracies, incorrectly attributing the film to the \textit{French
New Wave movement} and listing incorrect personnel. This highlights the
limitations of smaller LLMs fin complex reasoning tasks without specialized
knowledge. (3) COMP w/o GEP shows improved accuracy in genre reasoning but struggle
with factual accuracy about cast and crew, making plausible but unsupported
inferences. These comparisons emphasize the crucial roles that KG integration and
proper pre-training play in enabling COMPASS to perform more accurate reasoning over
user preferences, generating explainable summaries that enhance transparency and
trust in CRSs.
    \section{Conclusion}

In this paper, we introduce COMPASS, a novel framework that augments LLMs with KGs to reason over user preferences in CRSs.
To address the modality gap between structured knowledge and natural language, we propose a graph entity captioning that transforms KG structures into LLM-compatible representations.
Through knowledge-aware instruction tuning, COMPASS becomes proficient in performing cross-modal reasoning, generating interpretable user preference summaries.
COMPASS has been extensively evaluated as a plug-and-play enhancement for various CRSs across benchmark datasets.
The results demonstrate its effectiveness in significantly improving both recommendation performance and explainability of these base models. 
The adaptive integration mechanism of COMPASS allows for seamless enhancement of diverse CRS architectures without structural modifications, showcasing its versatility and potential for widespread adoption in the field of CRSs.

    \bibliographystyle{IEEEtran}
    \bibliography{zotera_sim}
\end{document}